\documentclass[conference]{IEEEtran}

\usepackage{cite}

\usepackage{hyperref}

\usepackage{adjustbox}
\usepackage[normalem]{ulem}
\useunder{\uline}{\ul}{}

\usepackage{supertabular, lscape}
\usepackage{multirow}
\usepackage{siunitx} 

\usepackage{lineno}

\usepackage[table]{xcolor}
\usepackage{colortbl}

\def\BibTeX{{\rm B\kern-.05em{\sc i\kern-.025em b}\kern-.08em
    T\kern-.1667em\lower.7ex\hbox{E}\kern-.125emX}}

\begin{document}

\nolinenumbers

\title{A Novel Framework for Selection of Generative Adversarial Networks for an Application}

\author{
    \IEEEauthorblockN{Tanya Motwani\IEEEauthorrefmark{1}, Manojkumar Parmar\IEEEauthorrefmark{1}\IEEEauthorrefmark{2}}
    \IEEEauthorblockA{\IEEEauthorrefmark{1}Robert Bosch Engineering and Business Solutions Private Limited, Bengaluru, India}
    \IEEEauthorblockA{\IEEEauthorrefmark{2}HEC Paris, Jouy-en-Josas Cedex, France}
}

\maketitle

\begin{abstract}
Generative Adversarial Network (GAN) is a current focal point of research. The body of knowledge is fragmented, leading to a trial-error method while selecting an appropriate GAN for a given scenario. We provide a comprehensive summary of the evolution of GANs starting from its inception addressing issues like mode collapse, vanishing gradient, unstable training and non-convergence. We also provide a comparison of various GANs from the application point of view, their behavior and implementation details. We propose a novel framework to identify candidate GANs for a specific use case based on architecture, loss, regularization and divergence. We also discuss application of the framework using an example, and we demonstrate a significant reduction in search space. This efficient way to determine potential GANs lowers unit economics of AI development for organizations.
\end{abstract}

\IEEEpeerreviewmaketitle

\section{Introduction} \label{intro}

Generative Adversarial Networks (GANs) are a category of generative models built upon game theory; a two-player minimax game \cite{pi_Ian_J._Goodfellow_2014}. A typical architecture of such a model consists of two neural networks – a discriminator and generator. The generator transforms the input noise vector into a potentially high dimensional data vector. The discriminator evaluates whether this vector is derived from the original distribution. Based on the outcome, the generator learns to produce samples that are similar to the original distribution. This adversarial technique holds that improvements in one component come at the expense of the other.
\newline
GANs are one of the dominant methods for generation of realistic and diverse examples in the domains of computer vision \cite{Wang_2018} \cite{Bousmalis_2017} \cite{1610.09585} \cite{Choi_2018}, time-series synthesis \cite{1706.02633} \cite{1806.01875} \cite{Li_2019} \cite{pi_Chris_Donahue_2018}, natural language processing \cite{1705.11001} \cite{1801.07736} \cite{Subramanian_2017} \cite{Haidar_2019}, etc. They belong to the class of implicit models which follow a likelihood-free inference approach \cite{1610.03483}. Implicit probabilistic models enjoy additional modelling flexibility as compared to classical probabilistic models \cite{1811.12402}. These models generate images sampled from the learned distribution and do not provide any latent representation of the data samples. GANs offer advantages such as parallel generation, universal approximation, better quality, sharp density estimations and understanding of the structural hierarchy of samples, over other explicit generative models. These properties have aided in immense popularity of GANs in the deep learning community, especially in the field of computer vision.
Despite their successes, GANs remain difficult to train as the nature of their optimization results in a dynamic system; each time any parameter of a component, either the discriminator or the generator, is modified, it results in the instability of the system. Current research is dedicated towards search for stable combinations of architectures, losses and hyperparameters for various applications such as image and video generation \cite{1810.02419} \cite{Tulyakov_2018} \cite{pi_Aidan_Clark_2019}, domain adaptation \cite{Bousmalis_2017} \cite{Sankaranarayanan_2018} \cite{1711.03213} \cite{Hong_2018}, speech synthesis \cite{Yang_2017} \cite{Saito_2018} \cite{Bollepalli_2017}, semantic photo editing \cite{Wang_2018} \cite{1609.07093} \cite{schonfeld_2020}, etc. While these models attain interesting results for particular applications, there is no thorough consensus or reference study available to understand which GAN performs better than others for a specific use case. In this paper, we aim to address the above supposition and narrow down the combinations of attributes for GANs through a technical framework.
\subsection{Article Structure}
The organization of the paper is as follows: Section \ref{training} highlights the concerns that have transpired while training GANs, followed by Section \ref{Evolution} that gives an outline of popular loss-variants of GANs. Section \ref{comaparision} presents a contrast between these GANs based on application, behavior and implementation. Section \ref{framework} defines the framework with the set of most commonly used architectures, loss functions, regularizations and divergence schemes. Section \ref{Example} explicates the use of the framework through an example. The future research scope is underlined in Section \ref{future}, followed by Section \ref{conclusion} as summary.

\section{Training Issues with Classic GANs} \label{training}
Despite their progress and success, GANs are subjected to a variety of difficulties during training. These mainly include mode collapse \cite{1611.02163}, optimization instability \cite{1701.04862}, vanishing gradient and non-convergence \cite{1412.6515}. Furthermore, the methods that attempt to solve these issues depend on heuristics that are susceptible to little modifications. This premise makes it difficult to experiment with new models or utilize the existing ones for different applications. A solid understanding with an emphasis on both their theoretical and practical perspectives is needed to curate research directions towards addressing them.

\subsection{Mode Collapse}
A probability distribution may be multimodal and consist of multiple peaks for various sub-graphs of sample data.  Mode collapse, a limiting case of GANs to model multimodal distribution, occurs when the generator places its probability density in a small area of data space. The generator focuses on the creation of new data, while the discriminator's objective is to evaluate it for authenticity but not for diversity of samples. Every update of the generator ends with over-optimization of the discriminator, which makes it too easy for the generator to search for the most plausible output in its next iteration. Consequently, the generator rotates through a small group of output types. The discriminator treats each sample independently, and thus, there is no mechanism that incentivizes the generator or the discriminator to produce sundry results. \cite{1902.03984} demonstrates that the original GAN objective also encourages gradient exploding in the discriminator which leads to mode collapse in the generator. Mode collapse results in a low-quality synthetic distribution. For example, in the case of animal classification, mode collapse would ensue in the generator learning different features and colors for dogs but limited for cats, ultimately, exhibiting poor diversity. 

\subsection{Vanishing Gradient}
Minimization of minimax GAN's objective function results in vanishing gradient, which makes it difficult to update the generator. When the source and target distributions are not perfectly aligned, the discriminator will be close to optimal and the gradient for the objective function of GAN will be zero almost everywhere. This supplies little feedback to the generator, slowly halting the learning. A popular solution for this hurdle is to use a parameterization of loss where gradients don't vanish rather than limiting the power of discriminator \cite{1701.00160}. An alternative cause of vanishing gradient is when real-world data is usually concentrated in lower-dimensional manifolds, making it extremely simple for the discriminator to classify samples as real and fake, and leading to random unlearned outputs.

\subsection{Unstable Training}
Gradient descent-based GAN optimization techniques do not necessarily lead to convergence, and therefore, it is critical to understand their training dynamics. The algorithm exhibits local behavior near the Nash-equilibrium \cite{pi_Weili_Nie_2018}, which can be randomly far from the global equilibrium point and fails to perform consistently with non-convex cost functions or in two-player non-cooperative surroundings. Even if the training losses of both discriminator and generator converge, it does not imply that P\textsubscript{g}= P\textsubscript{d} (P\textsubscript{g} denotes generator's probability distribution, and P\textsubscript{d} signifies that of discriminator) \cite{1710.08446}. It has been observed that these losses oscillate, showing that the training is highly unstable and ultimately, resulting in mode collapse \cite{pi_Sujit_Gujar_2019}. GANs also require meticulous refinement of hyperparameters. A large-scale study has indicated that fine-tuning hyperparameters gravitate to better results than the introduction of a new loss function \cite{pi_Karol_Kurach_2018}.

\subsection{Imbalance between Discriminator and Generator}
Without reaching the equilibrium, GANs progress from generating one type of sample to another type. When the generator reaches the equilibrium point, the discriminator's slope is the largest, and it pushes the generator away from the target distribution. Consequently, the generator advances towards the target distribution and the discriminator alters its slope from positive to negative. This process occurs repetitively and therefore, the loss plots produced during training don't indicate convergence \cite{Ma_2018}. In addition, the discriminator is frequently able to attain a higher classification accuracy before the generator has produced a high dimensional sample and therefore, it is needed to temper the discriminator's performance whenever necessary.  An imbalance between discriminator and generator ultimately leads to non-convergence – if the generator continues to train even when discriminator gives random feedback, the quality of images generated collapses. 

\section{Evolution of GANs} \label{Evolution}
Various flavors of GANs have been introduced that focus on modification of loss functions to address the training difficulties of GANs.  We provide a tabular summary (Table 1) and an evolution timeline of specific loss-variants that help improve the performance of GANs for a set of applications. The objective is to give a bird eye's view over these GANs, their contributions and proposed solutions. The first column enlists the year of the first paper's introduction; the next one gives the name of GAN followed by the column of experiments conducted for modification of loss, architecture and regularization based on the issues related to GAN (second column from the right). The fourth column points to the datasets used for experimentation and the final column specifies the metric used to assess the performance of the proposed GAN. We consider the following abbreviations: Batch Normalization (BN) \cite{1502.03167}, Convolutional (CON), Decoder (Dec), Deconvolutional (DECON), Discriminator (D), Encoder (Enc), Fully Connected network (FC), Generator (G), Layer Normalization (LN) \cite{1607.06450}, Multilayer Perceptron (MLP), Normalization (N), Optimizer (O).

\section{Comparison of various GANs} \label{comaparision}
For an in-depth analysis of the above-mentioned GANs, we provide a comparative assessment of their theoretical, behavioral and practical facets in the form of Table 2 to Table 7. Three parameters, namely, application, behavior and implementation, have been considered for comparison. For every table, the first column comprises the name of the GAN which is being compared with the GAN enlisted as the first row. The second column specifies the results of the experiments conducted during comparison, the third column differentiates on the basis of behavioral properties, and the last column dictates the details of the network implementation, excluding the architecture. The blank cells suggest that there are no significant similarities or differences between the models.

\onecolumn
\begin{landscape}
\begin{table*}
\caption{Summary of loss variants in GANs}
\label{table_1}
\renewcommand{\arraystretch}{1.4}{
\begin{tabular}{|p{9mm}|p{2cm}|p{3.4cm}|p{4.4cm}|p{3cm}|p{3.2cm}|p{3cm}|}
\hline
\rowcolor{lightgray}
\textbf{\centerline{YEAR}}  & \textbf{\centerline{NAME}} & \multicolumn{2}{c|}{\textbf{EXPERIMENTATION}} & \textbf{\centerline{DATASET}} & \textbf{\centerline{ISSUES ADDRESSED}} & \textbf{\centerline{PERFORMANCE METRIC}} \\ \hline
\rowcolor{white}
\multirow{3}{*}\hfill{2014} & \multirow{3}{*}\hfill{\textbf{SGAN {}\cite{pi_Ian_J._Goodfellow_2014}{}}} & \cellcolor{lightgray}\textbf{\centerline{LOSS}} & Minimax & \multirow{3}{*}\hfill{MNIST {}\cite{Lecun_1998}{}, TFD {}\cite{susskind_2010}{}, CIFAR-10 {}\cite{pi_Alex_Krizhevsky_2009}{}} & \multirow{3}{*}\hfill{} & \multirow{3}{*}\hfill{Log-likelihood under distribution (kernel density) of Gaussian Parzen window {}\cite{Breuleux_2011}{}} \\ \cline{3-4} \cline{3-4}
 &  & \cellcolor{lightgray}\textbf{\centerline{ARCHITECTURE}} & \begin{tabular}[c]{@{}m{4.4cm}@{}}D: FC, CON \newline G: FC, DECON\end{tabular} &  &  &  \\ \cline{3-4} \cline{3-4}
 &  & \cellcolor{lightgray}\textbf{\centerline{REGULARIZATION}} & NO &  &  &  \\ \hline 
\multirow{3}{*}\hfill{2015} & \multirow{3}{*}\hfill{\textbf{VAEGAN {}\cite{1512.09300}{}}} & \cellcolor{lightgray}\textbf{\centerline{LOSS}} & Combines loss of Variational Autoencoder and GAN, while sharing parameters between Dec and G & \multirow{3}{*}\hfill{CelebA {}\cite{liu_2015}{}, CIFAR-10, STL-10 {}\cite{pi_Adam_Coates_2011}{}, LFW {}\cite{pi_Gary_B._Huang_2008}{}} & \multirow{3}{*}\hfill{Mode Collapse} & \multirow{3}{*}\hfill{Visual perception (Qualitative Assessment)} \\ \cline{3-4} \cline{3-4}
 &  & \cellcolor{lightgray}\textbf{\centerline{ARCHITECTURE}} & \begin{tabular}[c]{@{}m{4.4cm}@{}}Enc, D: CON \newline Dec: DECON \newline O: RMSProp {}\cite{tieleman_2012}{}\\N: BN\end{tabular} &  &  &  \\ 
 \cline{3-4} \cline{3-4}
 &  & \cellcolor{lightgray}\textbf{\centerline{REGULARIZATION}} & NO &  &  &  \\ \hline
\multirow{4}{*}\hfill{2016} & \multirow{4}{*}\hfill{\textbf{F-GAN {}\cite{1801.04406}{}}} & \cellcolor{lightgray}\textbf{\centerline{LOSS}} & Introduces variational divergence estimation framework with generalization to f-divergences {}\cite{Nguyen_2010}{} {}\cite{0901.2698}{} \cite{Liese_2006} & \multirow{4}{*}\hfill{MNIST, LSUN {}\cite{1506.03365}{}} & \multirow{4}{*}{} \hfill& \multirow{4}{*}\hfill{Same as SGAN} \\ \cline{3-4} \cline{3-4}
 &  & \cellcolor{lightgray}\textbf{\centerline{ARCHITECTURE}} & \begin{tabular}[c]{@{}m{4.4cm}@{}}D, G: Based on DCGAN \newline O: Adam \cite{1412.6980} \newline N: BN \end{tabular} &  &  &  \\ \cline{3-4} \cline{3-4}
 &  & \cellcolor{lightgray}\textbf{\centerline{REGULARIZATION}} & NO &  &  &  \\ \hline
\multirow{3}{*}{2016} & \multirow{3}{*}\hfill{\textbf{LEAST SQUARES GAN {}\cite{Mao_2017}{}}} & \cellcolor{lightgray}\textbf{\centerline{LOSS}} & Adopt least squares function in SGAN's objective & \multirow{3}{*}\hfill{HWDB1.0 {}\cite{Liu_2013}{}, LSUN, Gaussian Mixtures {}\cite{1611.02163}{}} & \multirow{3}{*}\hfill{Stable Training, Vanishing Gradient, Mode Collapse} & \multirow{3}{*}\hfill{Same as VAEGAN} \\ \cline{3-4} \cline{3-4}
 &  & \cellcolor{lightgray}\textbf{\centerline{ARCHITECTURE}} & \begin{tabular}[c]{@{}m{4.4cm}@{}}D, G: Based on DCGAN \newline O: Adam, RMSProp \newline N: BN\end{tabular} &  &  &  \\ \cline{3-4} \cline{3-4}
 &  & \cellcolor{lightgray}\textbf{\centerline{REGULARIZATION}} & NO &  &  &  \\ \hline
\multirow{3}{*}{2017} & \multirow{3}{*}\hfill{\textbf{Loss Sensitive GAN {}\cite{Qi_2019}{}}} & \cellcolor{lightgray}\textbf{\centerline{LOSS}} & Objective function is based on minimization of data dependent margin (the loss of a real sample is less than that of generated sample). Formulated CLSGANa (fully and semi supervised), GLSGAN & \multirow{3}{*}\hfill{SVHN {}\cite{pi_Yuval_Netzer_2011}{}, CelebA, CIFAR-10, MNIST, tiny ImageNet} & \multirow{3}{*}\hfill{Vanishing Gradient, Generalizability across different data distributions} & \multirow{3}{*}\hfill{Visual perception, image classification for quantitative evaluation, MRE for generalizability} \\ \cline{3-4} \cline{3-4}
 &  & \cellcolor{lightgray}\textbf{\centerline{ARCHITECTURE}} & \begin{tabular}[c]{@{}m{4.4cm}@{}}D, G: Based on DCGAN \newline O: Adam \newline N: BN\end{tabular} &  &  &  \\ \cline{3-4} \cline{3-4}
 &  & \cellcolor{lightgray}\textbf{\centerline{REGULARIZATION}} & YES (Enforces Lipschitz regularity using loss margin) &  &  &  \\ \hline
 \end{tabular}%
}
\\
\end{table*}
\end{landscape}

\onecolumn
\begin{landscape}
\begin{table*}
\caption{Summary of loss variants in GANs}
\label{table1}
\renewcommand{\arraystretch}{1.4}{%
\begin{tabular}{|m{9mm}|m{2cm}|m{3.4cm}|m{4.4cm}|m{3cm}|m{3.2cm}|m{3cm}|}
\hline
\rowcolor{lightgray}
\textbf{\centerline{YEAR}} & \textbf{\centerline{NAME}} & \multicolumn{2}{c|}{\textbf{EXPERIMENTATION}} & \textbf{\centerline{DATASET}} & \textbf{\centerline{ISSUES ADDRESSED}} & \textbf{\centerline{PERFORMANCE METRIC}} 
\\ \hline
\rowcolor{white}
\multirow{3}{*}\hfill{2017} & \multirow{3}{*}{\vspace{10mm}\textbf{WGAN {}\cite{1701.07875}{}}} & \cellcolor{lightgray}\textbf{\centerline{LOSS}} & Objective based on EM Distance or Wasserstein-1 \cite{Csisz_r_2004} & \multirow{3}{*}\hfill{LSUN} & \multirow{3}{*}\hfill{Stable Training, Vanishing Gradient, Mode Collapse} & \multirow{3}{*}\hfill{Same as VAEGAN} \\ \cline{3-4} \cline{3-4}
 &  & \cellcolor{lightgray}\textbf{\centerline{ARCHITECTURE}} & \begin{tabular}[c]{@{}m{4.4cm}@{}}D: DCGAN \newline G: DCGAN, MLP \newline O: Adam, RMSProp \newline N: BN\end{tabular} &  &  &  \\ \cline{3-4} \cline{3-4}
 &  & \cellcolor{lightgray}\textbf{\centerline{REGULARIZATION}} & YES (Enforces Lipschitz regularity using weight clipping) &  &  &  \\ \hline
\multirow{3}{*}\hfill{2017} & \multirow{3}{*}\hfill{\textbf{GEOMETRIC GAN {}\cite{1705.02894}{}}} & \cellcolor{lightgray}\textbf{\centerline{LOSS}} & Geometric GAN with SVM {}\cite{Smola_2001}{} hyperplane, Geometric interpretations of SGAN, FGAN, WGAN, EBGAN {}\cite{1609.03126}{} & \multirow{3}{*}\hfill{MNIST, LSUN, CelebA, Gaussian mixture} & \multirow{3}{*}\hfill{Vanishing Gradient, Imbalance between D and G} & \multirow{3}{*}\hfill{Same as VAEGAN} \\ \cline{3-4} \cline{3-4}
 &  & \cellcolor{lightgray}\textbf{\centerline{ARCHITECTURE}} & \begin{tabular}[c]{@{}m{4.4cm}@{}}D, G: MLP, DCGAN \newline O: Adam, RMSProp \newline N: BN\end{tabular} &  &  &  \\ \cline{3-4} \cline{3-4}
 &  & \cellcolor{lightgray}\textbf{\centerline{REGULARIZATION}} & YES (Weight clipping, weight decay) &  &  &  \\ \hline
\multirow{3}{*}\hfill{2017} & \multirow{3}{*}\hfill{\textbf{WGAN-GP {}\cite{1704.00028}{}}} & \cellcolor{lightgray}\textbf{\centerline{LOSS}} & Wasserstein objective along with gradient penalty (GP) & \multirow{3}{*}\hfill{LSUN, CIFAR-10, Billion Word {}\cite{1312.3005}{}} & \multirow{3}{*}\hfill{Stable Training, Vanishing Gradient} & \multirow{3}{*}\hfill{Inception Score {}\cite{1606.03498}{}} \\ \cline{3-4} \cline{3-4}
 &  & \cellcolor{lightgray}\textbf{\centerline{ARCHITECTURE}} & \begin{tabular}[c]{@{}m{4.4cm}@{}}D, G: Based on DCGAN, ResNet {}\cite{He_2016}{}\\ G: MLP \newline O: Adam \newline N: BN, LN\end{tabular} &  &  &  \\ \cline{3-4} \cline{3-4}
 &  & \cellcolor{lightgray}\textbf{\centerline{REGULARIZATION}} & YES (GP with comparison to weight clipping) &  &  &  \\ \hline
\multirow{3}{*}\hfill{2018} & \multirow{3}{*}\hfill{\textbf{SN-GAN {}\cite{1802.05957}{}}} & \cellcolor{lightgray}\textbf{\centerline{LOSS}} & Optimized the objective of SGAN with normalization of weight matrices. Comparisons with Minimax, Hinge {}\cite{1702.08896}{}, WGAN-GP & \multirow{3}{*}\hfill{CIFAR-10, STL-10, ImageNet {}\cite{Deng_2009}{}} & \multirow{3}{*}\hfill{Stable Training} & \multirow{3}{*}\hfill{Inception Score, Frechet Inception Distance (FID) {}\cite{pi_Martin_Heusel_2017}{}} \\ \cline{3-4} \cline{3-4}
 &  & \cellcolor{lightgray}\textbf{\centerline{ARCHITECTURE}} & \begin{tabular}[c]{@{}m{4.4cm}@{}}D, G: CON, ResNet \newline O: Adam \newline N: BN, LN\end{tabular} &  &  &  \\ \cline{3-4} \cline{3-4}
 &  & \cellcolor{lightgray}\textbf{\centerline{REGULARIZATION}} & YES (Spectral Norm (SN) and comparison with other normalization techniques (GP, Weight normalization, and Orthonormal)) &  &  &  \\ \hline
\multirow{3}{*}\hfill{2018} & \multirow{3}{*}\hfill{\textbf{RGAN {}\cite{1807.00734}{}}} & \cellcolor{lightgray}\textbf{\centerline{LOSS}} & Relativistic SGAN such that G pushes D towards 0.5, rather than 1. Introduced Relativistic average loss & \multirow{3}{*}\hfill{CIFAR-10, CAT {}\cite{Zhang_2008}{}} & \multirow{3}{*}\hfill{Stable Training, Vanishing Gradient} & \multirow{3}{*}\hfill{Frechet Inception Distance} \\ \cline{3-4} \cline{3-4}
 &  & \cellcolor{lightgray}\textbf{\centerline{ARCHITECTURE}} & \begin{tabular}[c]{@{}m{4.4cm}@{}}D, G: DCGAN, CON (same as SNGAN) \newline O: Adam \newline N: BN\end{tabular} &  &  &  \\ 
 \cline{3-4} \cline{3-4}
 &  & \cellcolor{lightgray}\textbf{\centerline{REGULARIZATION}} & YES (Gradient Penalty, Spectral Norm) &  &  &  \\ \hline
\end{tabular}%
}
\\
\end{table*}
\end{landscape}

\onecolumn
\begin{table*}
\caption{Comparison of GANs with FCGAN}
\label{table_2}
\resizebox{\textwidth}{!}{
\rowcolors{3}{white!70!gray!20}{white!70!gray!130}{
\begin{tabular}{|m{2cm}|m{4cm}|m{5cm}|m{4cm}|}
\hline
\rowcolor{lightgray}
\textbf
{GAN} & \multicolumn{1}{c|}{\textbf{APPLICATION}} & \multicolumn{1}{c|}{\textbf{BEHAVIOUR}} & \multicolumn{1}{c|}{\textbf{IMPLEMENTATION}} \\* \hline
\multicolumn{4}{|c|}{\textbf{FCGAN}} \\* \hline
\center{\textbf
{VAEGAN {}\cite{1512.09300}{} {}\cite{1812.05676}{}}} & VAEGAN had lower entropy of each single image generated showing high quality output on synthetic MNIST distribution. & \begin{tabular}[c]{@{}p{5 cm}@{}} 1. VAEGAN does not suffer from mode collapse. It preserves the functionality to map a single image to its latent variables.\\ 2. VAEGAN implicitly encourages similarity between synthetic and training data unlike FCGAN.\end{tabular} & VAEGAN has the same loss function for the discriminator as FCGAN. \\* 
\hline
\center{\textbf
{Least Squares GAN (LSGAN) {}\cite{Mao_2017}{} {}\cite{Mao_2019}{}}} & \vspace{-15cm}  LSGAN observes successful learning on BNGenerator with Adam optimizer unlike FCGAN on LSUN data & \begin{tabular}[c]{@{}p{5 cm}@{}}1. LSGAN exhibits less mode seeking behavior and training instability compared to FCGAN. \\ 2. LSGAN penalizes samples which are on the correct side of the decision boundary but far away from the real data as log loss doesn't care about distance but only the sign. \\ 3. Least squares function (LSGAN) is flat at only one point unlike sigmoid cross entropy, not causing values to saturate when x is too large. LSGAN achieves good convergence without BN.\end{tabular} &  \vspace{-20cm} \begin{tabular}[c]{@{}p{4 cm}@{}} 1. LSGAN removed log from D in FCGAN and instead used L2 loss.\\ 2. RMSProp performs more stable than Adam optimizer for both the GANs.\end{tabular} \\* \hline
\center{\textbf
{WGAN}} & \vspace{-20mm} WGAN was able to create better quality of samples than FCGAN but lower than that of DCGAN for LSUN dataset. & \begin{tabular}[c]{@{}p{5 cm}@{}} 1. WGANs are more robust than FCGAN when one varies the architectural choices for the generator without any evidence of mode collapse. \\ 2. WGAN is trained optimally which makes it impossible to collapse modes and not limit to imperfect gradients.\\ 3. Training of WGAN with weight clipping is slower than that of original GAN.\\ 4. EM distance guarantees continuity and differentiability, which KL divergence and JS divergence lack.\\ 5. No balance between D and G, or careful network architecture is required in WGAN. \\  6. WGAN value function correlates with sample quality unlike FCGAN.\end{tabular} & \vspace{-15cm} \begin{tabular}[c]{@{}p{4 cm}@{}} 1. The last layer of sigmoid in D of FCGAN is removed in WGAN, as WGAN performs regression, not binary classification. \\ 2. WGAN recommends RMSProp rather than a momentum-based optimizer like Adam, as it causes instability in model training.\end{tabular} \\* 
\hline
\center{\textbf
{GEOMETRIC GAN {}\cite{1704.00767}{} {}\cite{1705.02894}{}}} & Geometric GAN demonstrated less mode collapse with Lipschitz continuity regularization on Gaussian mixture dataset. & \begin{tabular}[c]{@{}p{5 cm}@{}}1. Geometric GAN differs in the definition of normal vector of the separating plane.\\ 2. Geometric GAN successfully converges to Nash Equilibrium between D and G.\\ 3. Geometric GAN has a linear hyperplane consistent approach compared to non-linear separating hyperplane of FCGAN.\end{tabular} &  \\* 
\hline
\center{\textbf
{WGAN-GP}} & WGAN-GP achieves comparable sample quality to FCGAN objective for equivalent architectures on CIFAR-10 and LSUN datasets but has increased stability which is used to explore a range of architectures. &  WGAN-GP doesn't always converge to the equilibrium point with a finite number of updates of D per G update, unlike FCGAN which focuses training with consensus optimization, zero-centered gradient penalties or instance noise and therefore, converges when provided with enough capacity. & No BN in WGAN-GP because it changes D's function mapping from one input to output in whole batches while WGAN-GP penalizes the norm independently. \\* \hline
\center{\textbf{RSGAN}} & 
\begin{tabular}[c]{@{}p{4 cm}@{}} \vspace{-11mm}  1. RSGAN performed only slightly better on CIFAR-10. Claimed that the dataset was too easy to realize the stabilizing effects of RSGAN. \\ 
2. In CAT dataset with high resolution pictures, relativism showed more improvement than spectral norm or gradient penalty.\end{tabular}
& \begin{tabular}[c]{@{}p{5 cm}@{}} 1. RSGAN fixes the generator's objective in FCGAN such that it not only increases the probability of fake data being real but also decreases the probability of real data being real.\\ 2. RSGAN works very well in conjunction with gradient penalty, even when using only one D update per G update.\\ 3. Relativism significantly improves data quality and stability of GANs at no computational cost.\\ 4. FCGAN becomes stuck early in training as when the D reaches optimality, the gradient completely ignores real data. As RSGAN estimates the probability of real data being more realistic than a randomly-sampled fake data, both real and fake data will always be incorporated in the gradient of the D's loss function.\end{tabular} &  \\* \hline
\end{tabular} 
}
}
\\
\end{table*}

\begin{table*}
\caption{Comparison of GANs with WGAN}
\label{table_3}
\resizebox{\textwidth}{!}{
\rowcolors{3}{white!70!gray!20}{white!70!gray!130}{
\begin{tabular}{|m{2cm}|m{4cm}|m{5cm}|m{4cm}|}
\hline
\rowcolor{lightgray}
\textbf
{GAN} & \multicolumn{1}{c|}{\textbf{APPLICATION}} & \multicolumn{1}{c|}{\textbf{BEHAVIOUR}} & \multicolumn{1}{c|}{\textbf{IMPLEMENTATION}} \\* \hline
\multicolumn{4}{|c|}{\textbf{WGAN}} \\* \hline
\center{\textbf{WGAN-GP}} & \begin{tabular}[c]{@{}p{4 cm}@{}} 1. WGAN-GP significantly outperforms weight clipping by higher inception scores and convergence rates on CIFAR-10. \\ 2. Training loss of WGAN-GP on MNIST gradually increases even when the validation loss has dropped unlike WGAN.\end{tabular} & \begin{tabular}[c]{@{}p{5 cm}@{}} 1. WGAN fails to capture higher moments of the data distribution. They end up learning simple functions and models very simple approximations to the optimal functions. \\ 2. Detection of overfitting in WGAN-GP is faster compared to WGAN when given enough capacity and too little training capacity. \\ 3. The loss quality of both correlates with sample quality and converges toward a minimum. \end{tabular} & \begin{tabular}[c]{@{}p{4 cm}@{}} \vspace{-7mm}  1. WGAN-GP recommends LN as a replacement for BN. \\ 2. Adam performs better than RMSProp as an optimizer with its objective in WGAN-GP. \end{tabular} \\* \hline
\center{\textbf{SNGAN\cite{1705.10941}}} & SNGAN is relatively robust and produces diverse and complex images with aggressive learning and momentum rates compared to weight clipping, evaluated on CIFAR-10 and STL-10. & \begin{tabular}[c]{@{}p{5 cm}@{}} 1. In WGAN, the number of features is diminished by weight clipping leading to a random model that matches the target distribution at select features. SNGAN augments the cost function with a sample data dependent regularization term. \\ 2. Weight matrices of layers trained with weight clipping are rank deficit which proves to be fatal in lower layers (unnecessary restricts the search space of the discriminator) unlike those of SNGAN which are broadly distributed. \end{tabular} & \vspace{-7mm} In the absence of regularization techniques, SN provides better sample quality compared to weight normalization and gradient penalty. \\* \hline
\center{\textbf{Loss Sensitive GAN(LS-GAN)\cite{Edraki_2018}}} & 
\vspace{-15mm} GLS-GAN attained a smaller MRE on tiny ImageNet dataset compared to WGAN. & \begin{tabular}[c]{@{}p{5 cm}@{}} 
1. Introduced generalized LS-GAN (GLS-GAN). GLS-GAN contains a large family of regularized GANs which contain both LS-GAN and Wasserstein as its special cases. \\2. WGAN seeks to maximize the mean under the densities of real and generated samples, and clips the network weights on a bounded box to prevent the loss function from becoming unbounded. LS-GAN treats real and generated samples in pairs and maximizes the difference in their losses up to a data-dependent margin, which not only prevents their losses from being decomposed into two separate first order moments but also enforces them to coordinate with each other to learn the optimal loss function.  \end{tabular}
& \begin{tabular}[c]{@{}p{4 cm}@{}} \vspace{-8mm} 1. Both can produce unclasped natural samples without using BN and address vanishing gradient while training G.\\ 2. Both use weight regularization as a means of ensuring the model function has a bounded Lipschitz constant. WGANs use weight clipping, LS-GANs use weight decay. \end{tabular} \\* \hline
\center{\textbf{Geometric GAN}} & \vspace{-15mm} No comparative study available with respect to application. & \begin{tabular}[c]{@{}p{5 cm}@{}} 1. WGAN follows a mean-difference driven approach and leads to generation of mean of arbitrary number of modes in true distributions. \\ 2. Geometric GAN follows a linear separating hyperplane, and shows robust convergence behavior. \end{tabular} & \\* \hline
\center{\textbf{Least Squares GAN}} & \vspace{-15mm} No comparative study available with respect to application & Though having a similar setup compared to WGAN, Least Squares GAN minimizes Pearson $ \chi^{2} $ divergence and learns a L2 loss function instead of critic function. & Least Squares GAN also uses regression, and therefore sigmoid layer is removed compared to FCGAN. \\* \hline
\center{\textbf{VAEGAN}} & WGAN generated images that did not belong to any of the 10 classes in MNIST dataset. This indicates that an application that cannot handle out-of-dataset images cannot utilize WGAN. & \begin{tabular}[c]{@{}p{5 cm}@{}} 1. GAN varies smoothly even when two distributions overlap but the generated samples are not realistic. VAEGAN implicitly encourages similarity between training and synthetic data. \\ 2. Both the models successfully address mode collapse. \\ 3. Entropy of each generated image is higher in WGAN because of its low quality (suspected that the FC network is not powerful enough). \end{tabular} & \\* \hline
\end{tabular}
}
}
\\
\end{table*}

\begin{table*}
\caption{Comparison of Loss Sensitive GAN with DCGAN}
\label{table_4}
\resizebox{\textwidth}{!}{
\rowcolors{3}{white!70!gray!20}{white!70!gray!130}{
\begin{tabular}{|m{2cm}|m{4cm}|m{5cm}|m{4cm}|}
\hline
\rowcolor{lightgray}
\textbf
{GAN} & \multicolumn{1}{c|}{\textbf{APPLICATION}} & \multicolumn{1}{c|}{\textbf{BEHAVIOUR}} & \multicolumn{1}{c|}{\textbf{IMPLEMENTATION}} \\* \hline
\multicolumn{4}{|c|}{\textbf{DCGAN}} \\* \hline
\center{\textbf
{Loss Sensitive GAN (LS-GAN)}} & \begin{tabular}[c]{@{}p{4 cm}@{}} 1. LS-GAN outperformed DCGAN on classification of CIFAR-10 and SVHN datasets by a higher accuracy and lower error rate respectively. \\ 2. Regularized models such as LS-GAN have better generalization performances and more stable training while achieving a low MRE on CIFAR-10 dataset. \end{tabular} & \begin{tabular}[c]{@{}p{5 cm}@{}} 1. LS-GAN’s loss comprises of linear constraints and objective, contrary to log loss which causes vanishing gradient. The linear gradient, rather than being saturated, provides sufficient gradient to continuously update the generator. \\ 2. LS-GAN is not affected by over-trained loss function, unlike DCGAN. \end{tabular} & \begin{tabular}[c]{@{}p{4 cm}@{}} 1. LS-GAN does not use a sigmoid layer as the output of the loss function. \\ 2. BN is known to prevent mode collapse in DCGAN; without BN, DCGAN cannot produce any images and would collapse. LS-GAN proves to be more resilient with different structure changes and performs very well even if BN layers are removed. \end{tabular} \\* \hline
\end{tabular}
}
}
\\
\vspace{2em}

\caption{Comparison of RSGAN with Least Squares GAN}
\label{table_5}
\resizebox{\textwidth}{!}{
\rowcolors{3}{white!70!gray!20}{white!70!gray!130}{
\begin{tabular}{|m{2cm}|m{4cm}|m{5cm}|m{4cm}|}
\hline
\rowcolor{lightgray}
\textbf
{GAN} & \multicolumn{1}{c|}{\textbf{APPLICATION}} & \multicolumn{1}{c|}{\textbf{BEHAVIOUR}} & \multicolumn{1}{c|}{\textbf{IMPLEMENTATION}} \\* \hline
\multicolumn{4}{|c|}{\textbf{Least Squares GAN}} \\* \hline
\center{\textbf
{RSGAN}} & \begin{tabular}[c]{@{}p{4 cm}@{}} 1. The combination of LSGAN with relativism (RaLSGAN) performed better than simple LSGAN in both unstable and stable setups, evaluated using FID on CIFAR-10. \\ 2. LSGAN produced high quality 64*64 resolution CAT images (low FID score) but produced them in a very unstable manner. \end{tabular} & LSGAN is unable to converge in high resolution (256*256 or more) image dataset while RSGAN can generate images in all resolutions. & \\* 
\hline
\end{tabular} 
}
}
\\
\vspace{2em}

\caption{Comparison of GANs with WGAN-GP}
\label{table_7}
\resizebox{\textwidth}{!}{
\rowcolors{3}{white!70!gray!20}{white!70!gray!130}{
\begin{tabular}{|m{2cm}|m{4cm}|m{5cm}|m{4cm}|}
\hline
\rowcolor{lightgray}
\textbf
{GAN} & \multicolumn{1}{c|}{\textbf{APPLICATION}} & \multicolumn{1}{c|}{\textbf{BEHAVIOUR}} & \multicolumn{1}{c|}{\textbf{IMPLEMENTATION}} \\* \hline
\multicolumn{4}{|c|}{\textbf{WGAN-GP}} \\* \hline
\center{\textbf
{Least Squares GAN}} & \begin{tabular}[c]{@{}p{4 cm}@{}} 1. LSGANs perform better than WGAN-GP on datasets of LSUN, CAT and CIFAR-10 datasets while performed poorly on ImageNet. \\ 2. LSGAN and WGAN-GP achieve similar FID on LSUN but LSGAN much less time to reach the optimal FID. \\ 3. Both LSGAN-GP and WGAN-GP succeed in training difficult architectures and generate higher quality images with 101-layer ResNet. \end{tabular} & \begin{tabular}[c]{@{}p{5 cm}@{}} \vspace{-10mm} WGAN-GP is more computational intensive than LSGAN-GP and requires multiple updates for discriminator. \end{tabular} & \\* \hline
\center{\textbf{SNGAN}} & 
\begin{tabular}[c]{@{}p{4 cm}@{}} 1. WGAN-GP fails to train GANs at high momentum and learning rates on both STL-10 and CIFAR-10. \\ 
2. The combination of WGAN-GP and parameterization with spectral normalization achieves better quality images than WGAN-GP.\end{tabular}
& \begin{tabular}[c]{@{}p{5 cm}@{}} 1. WGAN-GP heavily depends on the support of current generative distribution. \\ 2. As they change over the course of training, they destabilize the effect of GP. \\ 3. SN can be used with GP (local regularizers), because it provides global regularization on the D. \end{tabular} & \begin{tabular}[c]{@{}p{4 cm}@{}} \vspace{-5mm} SNGAN requires less computational cost compared to WGAN-GP. \end{tabular} \\* \hline
\center{\textbf
{RSGAN}} & \vspace{-5mm} WGAN-GP produces high quality images (a low FID score) on stable setup in CIFAR-10 compared to RSGAN. & \begin{tabular}[c]{@{}p{5 cm}@{}} 1.As WGAN-GP is an integral probability metric GAN, both the real and fake data equally contribute to the gradient of D's loss function. They implicitly assume that some of the samples are fake, similar to the function of relativism. \\ 2. In unstable setups, WGAN-GP performed very poorly because of a single discriminator update per generator update. Relativism provides a greater improvement in difficult settings compared to gradient penalty.\end{tabular} & \\* 
\hline
\end{tabular}
}
}
\\
\end{table*}

\twocolumn

\begin{table*}
\caption{Comparison of F-divergence GAN with Geometric GAN}
\label{table_6}
\resizebox{\textwidth}{!}{
\rowcolors{3}{white!70!gray!20}{white!70!gray!130}{
\begin{tabular}{|m{2cm}|m{4cm}|m{5cm}|m{4cm}|}
\hline
\rowcolor{lightgray}
\textbf
{GAN} & \multicolumn{1}{c|}{\textbf{APPLICATION}} & \multicolumn{1}{c|}{\textbf{BEHAVIOUR}} & \multicolumn{1}{c|}{\textbf{IMPLEMENTATION}} \\* \hline
\multicolumn{4}{|c|}{\textbf{Geometric GAN}} \\* \hline
\center{\textbf
{F-divergence GAN}} & No comparative study available with respect to application. & In F-GAN, as the scaling factors that reflect geometric space are asymmetric, it is difficult to control the balance between D and G updates. & \\* \hline
\end{tabular} 
}
}
\\
\end{table*}

\section{The Framework} \label{framework}
Selection of the GAN model for a particular application is a combinatorial exploding problem with several possible choices and their orderings. It is computationally impossible for researchers to explore the entire space. Furthermore, there exists no standard evaluation metric for these networks that can provide a fair and neutral comparison. Even if a metric is determined, variations in architecture, losses, regularizations and hyperparameters would lead to different values of the metric \cite{pi_Karol_Kurach_2018}. There is a need for a standard framework that can be referred to compare GANs and their behavior. We propose a systematic substructure that consists of four decision parameters namely, architecture, loss, regularization and divergence, for reducing the number of possible configurations and selecting the most suitable GAN for a given use case. Figure 1 gives the principal loss and architecture GANs, regularization and divergence functions that have been introduced for improvements in GAN training since the inception of classic GAN. In this paper, we focus on the loss GAN variants, their original implementations and properties.

 \begin{figure*}
 \centering
\includegraphics[scale=0.5]{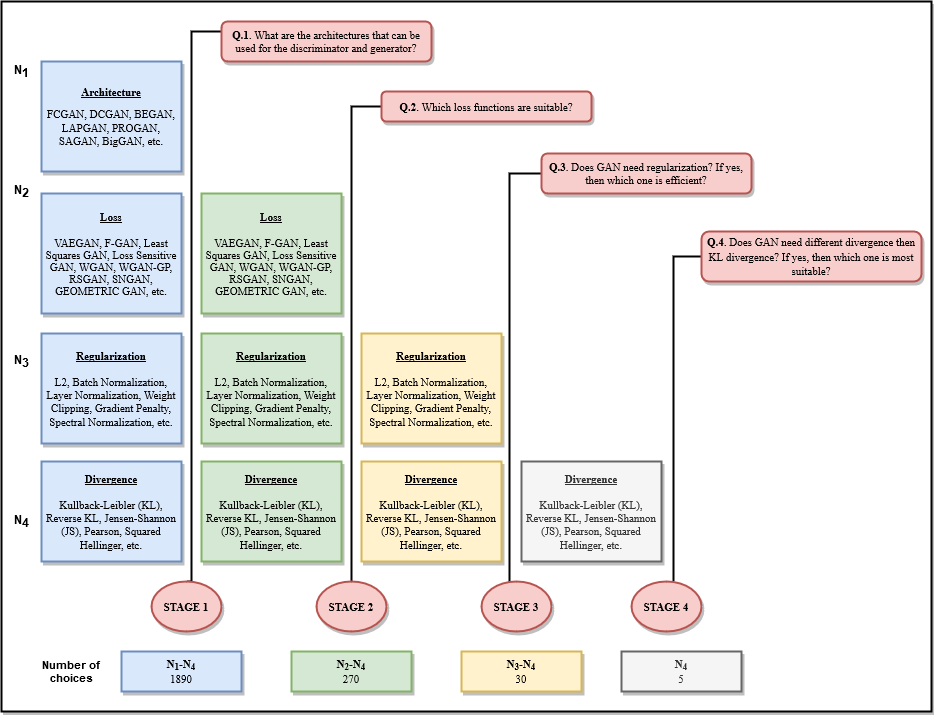}
\caption{The Proposed Framework}
\label{fig_1}
\end{figure*}

\section{An Example of Use of Framework} \label{Example}
Let’s take a case of image generation using CIFAR-10 dataset as an illustration of the framework. Consider that the application demands a good sample quality and diversity. Without a logical framework, one has to search within an exploding combinatorial space. Our framework helps provide few candidates by systematically eliminating other combinations. For example, we have nearly 5000 potential GAN functional combinations based on the available architectures, losses, divergences, etc. for this specific application. With help of this framework, we can narrow them down to 5-6 candidate GANs. This is equivalent to a 1000x reduction in the search space. 
\newline
To reduce the combinatorial search space, we ask the following 4 questions whose answers are derived based on Tables 1-7.
\begin{itemize}
   \item What are the architecture to be used for the discriminator and the generator? 
   \begin{itemize}
       \item Based on Table 1, the probable alternatives of architectures include fully connected, convolutional-deconvolutional networks or modifications of DCGAN. 
   \end{itemize}
   \item Which loss functions are suitable?
   \begin{itemize}
       \item The comparative assessment of loss GANs through the aspects of application, implementation and behavior in the form of Tables 2 - 7 provide a detailed study of loss GANs and their efficiency on image generation using CIFAR-10. 
       \item As the application requires high sample diversity and quality, the study suggests WGAN-GP, Least Squares GAN, RSGAN and SNGAN models. The combination of least squares GAN with relativism produces higher quality images compared to the independent models. Regularized models such as Loss Sensitive GAN and SNGAN demonstrate better generalization across distributions. 
   \end{itemize}
   \item Does GAN need regularization? If yes, then which one is efficient? 
   \begin{itemize}
       \item Our study indicates gradient penalty enhances the quality of images but does not stabilize the training. Spectral normalization indicates to be more computationally efficient compared to gradient penalty. \cite{pi_Karol_Kurach_2018}  showed that batch normalization in generator improves model quality while in discriminator manifested poor results. \cite{Zhang_2019} recommends regularization through augmentation of input/feature space to control the behaviour of the discriminator and thus, improve overall training.
   \end{itemize}
   \item 	Does GAN need different divergence then KL divergence? If yes, then which one is most suitable?
   \begin{itemize}
        \item \cite{1606.00709} introduced and experimented with various divergences including GAN, Kullback-Leibler and Squared-Hellinger, producing equally realistic samples. \cite{Roth_2017} introduces f-gan regularization that results in efficient divergence minimization and better data generalization. 
   \end{itemize}
\end{itemize}

\section{Future Work} \label{future}
Even if there have been recent improvements, there are still various open research problems for GANs. As a result of this detailed study, we pin down the issues related to the non-determinism of GAN training and propose definite actions to debunk future research directions. First, this body of knowledge can be converted into an automated tool that would promote easy accessibility. Next, similar to our study of loss variants in GANs, there is a need to address the architectural variants and their inter-comparisons to evaluate the best combination of architecture, optimizer and normalization. Development of quantitative evaluation metrics is another critical research direction as there exists no inherent estimate to realize the similarity between the source and target distributions \cite{Borji_2019}. \cite{1901.08753} demonstrate that there is no single winning regularization approach for GANs across all different settings. Further, hyperparameter optimization is still expensive in terms of computation: one can investigate and provide a detailed study on combinations of hyperparameter settings, the sensitivity of objective functions and regularization approaches with respect to hyperparameters and their refinements. This would aid in systematic experimentation of GAN and neutral model comparison. Moreover, a unit economics study in terms of computational cost can be executed to understand the performance of models and facilitate further research scope.
\section{Summary} \label{conclusion}
We discuss the issues and evolution of GANs, analyze the available loss variants of GANs. We provide a structured framework to determine the possible combinations of architecture, loss, regularization and divergence for selection of GAN for a use-case. When one needs to design a GAN for a specific application, our framework can be used as a baseline along with open-source reference implementations. We also present an in-depth comparative study between these variants on the basis of the application, implementation and behaviour. The usefulness of the framework is demonstrated through an example of image generation using CIFAR-10 dataset, where the framework successfully reduces 98\% of the number of combinations. This abates the overall computational cost of the GAN development for an application in organizations and promotes efficient use of resources.

\bibliographystyle{IEEEtran}
\bibliography{IEEEabrv, references}

\end{document}